\documentclass[11pt]{article}
\providecommand{\aclmode}{preprint}
\expandafter\usepackage\expandafter[\aclmode]{acl}
\makeatletter
\newif\ifreview
\ifacl@anonymize\reviewtrue\fi
\makeatother
\usepackage{times}
\usepackage{latexsym}
\usepackage[T1]{fontenc}
\usepackage[utf8]{inputenc}
\usepackage{microtype}
\usepackage{inconsolata}
\usepackage{graphicx}
\usepackage{booktabs}
\usepackage{amsmath}
\usepackage{xcolor}

\definecolor{cExr}{HTML}{0072B2}
\definecolor{cJson}{HTML}{56B4E9}
\definecolor{cExrShort}{HTML}{009E73}
\definecolor{cTagged}{HTML}{E69F00}
\definecolor{cDsl}{HTML}{D55E00}
\definecolor{cWords}{HTML}{000000}
\definecolor{cInk}{HTML}{22303F}
\definecolor{cMuted}{HTML}{7A7A7A}
\definecolor{cRule}{HTML}{D7DCE2}
\definecolor{cCorrect}{HTML}{E8E8E8}
\definecolor{cSilent}{HTML}{D55E00}

\newcommand{\LappliedMax}{78}
\newcommand{\LappliedMin}{72}

\newcommand{\LdslTok}{11.4}

\newcommand{\LexrExrShortGapAbs}{3.2}

\newcommand{\LexrS}{17.9}

\newcommand{\LfixMax}{62.6}
\newcommand{\LfixMin}{49.2}
\newcommand{\LhallRatio}{2.4}

\newcommand{\LnegMax}{98.3}
\newcommand{\LnegMin}{97.2}

\newcommand{\LoovRefMax}{100}
\newcommand{\LoovRefMin}{97}
\newcommand{\LotherNegErrMax}{28}
\newcommand{\LotherNegErrMin}{14}

\newcommand{\LrefuseMax}{10.1}
\newcommand{\LrefuseMin}{6.3}
\newcommand{\Lrerun}{4.9}
\newcommand{\LrerunSeedSd}{3.0}
\newcommand{\LsilentRatio}{1.8}
\newcommand{\LsilentRatioHi}{2.3}
\newcommand{\LsilentRatioLo}{1.4}
\newcommand{\Lspeedup}{4.4}

\newcommand{\LtiedS}{\fmt{DSL}}

\newcommand{\LwordsDslGap}{+5.4}

\newcommand{\LwordsExrGap}{+3.4}

\newcommand{\LwordsExrGapSmall}{+7.2}

\newcommand{\LwordsNegErr}{6}

\newcommand{\LwordsS}{10.0}
\newcommand{\LwordsTok}{12.0}

\newcommand{\QappliedMax}{88}
\newcommand{\QappliedMin}{80}

\newcommand{\QcueGapMax}{9}
\newcommand{\QcueGapMin}{5}

\newcommand{\QexrEm}{85.8}

\newcommand{\QexrExrShortGapAbs}{10.6}

\newcommand{\QexrJsonSep}{3}

\newcommand{\QexrNegRate}{2.0}
\newcommand{\QexrS}{10.3}

\newcommand{\QexrTok}{44.3}

\newcommand{\QfixMax}{75.3}
\newcommand{\QfixMin}{62.9}
\newcommand{\QhallRatio}{5.5}

\newcommand{\QlatRsq}{1.00}

\newcommand{\QmemMax}{1{,}806}
\newcommand{\QmemMin}{1{,}704}
\newcommand{\QmsFixed}{29}
\newcommand{\QmsPerTok}{26.5}
\newcommand{\QnegMax}{99.0}
\newcommand{\QnegMin}{95.9}
\newcommand{\QoovAns}{81}
\newcommand{\QoovRefMax}{100}
\newcommand{\QoovRefMin}{96}
\newcommand{\QotherNegErrMax}{34}
\newcommand{\QotherNegErrMin}{19}

\newcommand{\QrefuseMax}{5.4}
\newcommand{\QrefuseMin}{2.1}
\newcommand{\Qrerun}{1.5}
\newcommand{\QrerunSeedSd}{1.9}
\newcommand{\QsilentRatio}{2.3}
\newcommand{\QsilentRatioHi}{3.1}
\newcommand{\QsilentRatioLo}{1.8}
\newcommand{\Qspeedup}{3.7}

\newcommand{\QtiedEm}{\fmt{EXR}, \fmt{TAGGED}}

\newcommand{\QtokRatio}{4.1}

\newcommand{\QunkMax}{96}
\newcommand{\QunkMin}{86}

\newcommand{\QwordsDslGap}{+7.8}

\newcommand{\QwordsDslGapSmall}{+6.7}
\newcommand{\QwordsDslHi}{+9.5}
\newcommand{\QwordsDslLo}{+6.1}

\newcommand{\QwordsEm}{87.5}

\newcommand{\QwordsExrGap}{+1.8}

\newcommand{\QwordsExrGapSmall}{+4.0}

\newcommand{\QwordsLineErr}{36}

\newcommand{\QwordsNegErr}{5}
\newcommand{\QwordsNegRate}{0.2}
\newcommand{\QwordsNotPolite}{2.6}
\newcommand{\QwordsPolite}{6.1}
\newcommand{\QwordsS}{4.5}
\newcommand{\QwordsTok}{10.9}

\newcommand{\capHits}{19}
\newcommand{\corFrac}{10}
\newcommand{\cueBound}{2.4}
\newcommand{\cueRows}{32}
\newcommand{\lateRuns}{8}
\newcommand{\lfmGgufAgree}{93}
\newcommand{\maxSteps}{4{,}000}

\newcommand{\nBoot}{10{,}000}
\newcommand{\nCor}{600}
\newcommand{\nDev}{348}
\newcommand{\nGenerations}{511{,}500}
\newcommand{\nGrid}{300}
\newcommand{\nHead}{2{,}456{,}443}
\newcommand{\nNeg}{614}
\newcommand{\nPolite}{289}
\newcommand{\nSeeds}{5}
\newcommand{\nSizes}{5}

\newcommand{\nTest}{1{,}357}
\newcommand{\negFrac}{30}
\newcommand{\pairCells}{10}

\newcommand{\probeFixed}{60}
\newcommand{\probeRuns}{60}
\newcommand{\publishedEm}{78.6}
\newcommand{\publishedGap}{3}
\newcommand{\publishedSystem}{BART-Large EXR}
\newcommand{\publishedTrainN}{2{,}456{,}446}
\newcommand{\samplesSeen}{64{,}000}
\newcommand{\slopeCells}{10}
\newcommand{\slopeMax}{1.1}
\newcommand{\slopePos}{10}
\newcommand{\slopeSep}{1}
\newcommand{\wordsDslAheadAll}{10}
\newcommand{\wordsDslSepAll}{9}
\newcommand{\nSweepSeeds}{3}

\newcommand{\chDsl}{34.9}

\newcommand{\chWords}{43.1}

\newcommand{\exUtterance}{i want a pie with pepperoni and extra cheese}

\newcommand{\nTokTest}{1{,}357}
\newcommand{\tlDsl}{11.7}

\newcommand{\tlJson}{64.7}

\newcommand{\tlWords}{11.9}
\newcommand{\tqDsl}{10.8}

\newcommand{\tqJson}{45.6}

\newcommand{\tqWords}{10.3}

\newcommand{\padDevice}{iPad15,7}
\newcommand{\padLatency}{167\,ms}
\newcommand{\padPninety}{354\,ms}
\newcommand{\padLoad}{2.5\,s}
\newcommand{\padMem}{350\,MiB}
\newcommand{\padTps}{16\,tok/s}

\newcommand{\fmt}[1]{\textsc{\lowercase{#1}}}
\newcommand{\code}[1]{\texttt{#1}}
\newcommand{\Qlab}{Qwen3.5-0.8B}
\newcommand{\Llab}{LFM2.5-350M}
\newcommand{\demo}{\ifreview\citet{demoanon}\else\citet{demo}\fi}
\newcommand{\artifacts}{\ifreview the supplementary material\else
\url{https://github.com/UmarGit/coffee-llm}\fi}

\title{Parsed, Plausible, Wrong: Output Format Controls Silent Errors\\in Small-Model Semantic Parsing}

\author{
Umar Ahmed \and Maxim Dunaev \and Mohammad Nour Al Awad \and Sergey Ivanov \and Ivan Khodnenko \\
ITMO University \\
\texttt{\{uakhmed,mvdunaev,svivanov\}@itmo.ru}, \texttt{MohammadNourAlAwad@itmo.ru}, \texttt{ivan.khodnenko@itmo.ru}
}

\begin{document}
\maketitle

\begin{abstract}
A small fine-tuned parser that emits a well-formed but wrong tree fails
\emph{silently}: nothing downstream can tell. We ask whether the output format,
a free design choice, controls how often this happens. On the PIZZA ordering
benchmark we fine-tune two sub-billion models (\Qlab{} and \Llab{}) with LoRA on
six lossless encodings of the same target tree, from \QexrTok{} down to
\QwordsTok{} tokens per order, including two matched-length pairs that vary
surface form at fixed length. Across \nGrid{} runs (\nSizes{} training pool sizes,
\nSeeds{} seeds) with a hierarchical bootstrap and Holm correction fixed before
the results were read, shortening alone never helps, but writing the target in
ordinary words does. The word encoding is the most accurate format on both models
(\QwordsEm{}\% vs.\ \QexrEm{}\% exact match for the native s-expression on Qwen)
while emitting \QtokRatio{}$\times$ fewer tokens, and it cuts the silent error
rate by \QsilentRatio{}$\times$ on Qwen and \LsilentRatio{}$\times$ on LFM2,
separating at every pool size. Error analysis locates much of the gain in
negation. All runs also learn to refuse off-menu input (held out: \QnegMin{}--\QnegMax{}\%
handled) and to apply in-utterance self-corrections. Strict refusal has a measured
price: \QrefuseMin{}--\LrefuseMax{}\% of real, mostly polite, orders are refused.
\end{abstract}

\section{Introduction}

Small language models are attractive for narrow parsing tasks: they fine-tune in
minutes, run on a laptop or a tablet, and cost almost nothing per call. Their
weakness is reliability~\citep{hallucination}, and in structured prediction an
error comes in two kinds. An output that does not parse, or that the model marks
as a refusal, is \emph{detectable}: the caller rejects it and asks again. An
output that parses into a plausible tree that is not what the customer said is
\emph{silent}: the order goes through, wrong. This paper treats the silent error
rate $S$ as the quantity a deployment should minimise, and reports exact match
$A$ beside it for comparison with prior work (Section~\ref{sec:metrics}).

The work began with a deployed café ordering parser whose compact output format
was chosen for speed; its small hand-written test set could not tell formats
apart, so we moved to a public benchmark with a human-written test set.
Our question is whether the target format is itself a lever on silent errors.
The same tree can be written as an s-expression, as JSON, as \code{key=value}
pairs, as a terse symbolic language or as plain words, and the choice changes how
many tokens the model emits and how much of its pretraining it can reuse. The
intuitive hypothesis, that fewer output tokens leave less room for error, turns
out to be wrong in the form we first stated it. Shortening helps only when the
surface stays in natural words.

\paragraph{Contributions.}
(1)~A ladder of six lossless encodings of the PIZZA target tree~\citep{pizza},
with two matched-length pairs that separate length from surface form
(Section~\ref{sec:formats}).
(2)~Evidence across two model families, five pool sizes and five seeds that
length alone is not the mechanism: the word encoding beats its symbolic twin of
equal length in \wordsDslAheadAll{} of \pairCells{} model$\times$size
cells, separating in \wordsDslSepAll{} after Holm correction, and on
LFM2 it wins while costing \emph{more} tokens (Section~\ref{sec:form}).
(3)~The silent error rate as the deployment objective, inside a four-way split of
every test set. The word encoding has the lowest $S$ on both models, and an error
analysis shows where the difference comes from (Section~\ref{sec:silent}).
(4)~A guardrail trained into every run, refusing off-menu input and applying
spoken self-corrections, with its benefit and its cost measured on held-out data
(Section~\ref{sec:guard}).
\ifreview\else
Code, analysis scripts and the trained models are released at \artifacts.
\fi

\section{Related Work}

\paragraph{Target representation.}
That the meaning representation affects a parser is long established:
\citet{donglapata} decode logical forms as trees rather than sequences, and
seq2seq parsers for task-oriented dialogue generate the target directly
\citep{rongali}. Closest to us, \citet{shin} show that a frozen language model
prompted to produce a canonical natural-language form, mapped to the
representation by a grammar, beats prompting for the representation itself. We
find the same phenomenon under fine-tuning, where the model is free to learn any
notation, and we separate naturalness from length with matched pairs and two
vocabularies.

\paragraph{Format restrictions and constrained decoding.}
\citet{tam} find that forcing an LLM into JSON costs reasoning accuracy. Constrained
decoding makes invalid output impossible \citep{picard,gcd,outlines}; it can
remove our unparseable outputs, but it cannot make a well-formed tree correct, so
the silent rate is the part of the problem it does not reach.

\paragraph{Abstention and self-correction.}
Knowing when not to answer has been studied for reading comprehension
\citep{squad2} and under domain shift \citep{selective}. Spoken self-repair is a
classic topic in disfluency research \citep{shriberg}, and Disfl-QA shows that
disfluent input breaks QA models \citep{disflqa}. Task-oriented parsing datasets
such as TOP, TOPv2 and MTOP \citep{top,topv2,mtop}, and PIZZA itself, contain
neither off-domain refusals nor self-corrections; we add both as synthetic
training data and evaluate them on held-out vocabulary and cue phrases.

\paragraph{Tokenization and evaluation.}
Subword segmentation \citep{bpe} sets what a string costs, and vocabulary fit moves
accuracy \citep{tokenizer}. We follow recommendations to report seed variance and
significance \citep{dror,bouthillier}, using a bootstrap \citep{efron} with Holm's
correction \citep{holm}.

\paragraph{Deployment.}
\demo{} run the \fmt{WORDS} model studied here inside a web browser and check
that the deployed model reproduces its server evaluation; that work is about the
web system and reports no comparison between formats.

\section{Task and Data}
\label{sec:data}

PIZZA~\citep{pizza} maps spoken pizza and drink orders to trees with an unordered
exact-match criterion. Its training pool of \publishedTrainN{} utterances is
generated from a grammar; the \nDev{}-utterance development set and the
\nTest{}-utterance test set are written by people, which makes the test set the
strongest evidence in this paper. The pool contains \publishedGap{} rows identical
to a development or test utterance. We drop them from training, and verify on
every training subset we draw, at every size and seed, that no evaluation
utterance remains, exactly or after normalisation. The splits are used as
published.

\paragraph{Refusals.}
PIZZA has no notion of ``not an order'': a model trained on it turns
\emph{``hehe hehe''} into a pizza. We add generated negatives making up
\negFrac{}\% of every training set: chatter, off-menu requests, gibberish, long
text, and real orders with an off-menu clause or an unknown word inserted. The
rule is strict: anything that is not purely an order is refused whole. Each format
has one short refusal string that cannot parse as an order in that format
(\code{?} for the three word-like formats, \code{(REJECT )}, \code{(R)} and
\code{\{"reject":true\}} for the others), so all formats train on identical inputs.
Vocabulary is split: the \nNeg{} evaluation negatives use words never seen in
training.

\paragraph{Self-corrections.}
PIZZA has no self-corrections either, so \emph{``a pizza with cheese, wait no
cheese, only tomato sauce''} comes out with both cheese and no cheese. We take
real training orders, edit their tree, and append the spoken correction in nine
kinds (swap, retract, only, size, number, drink, cancel, restart, and decoys in
which a cue word changes nothing), adding \corFrac{}\% of the real orders. A
quarter of the cue phrases are held out: the \nCor{} evaluation corrections are
test orders with held-out cues. A fixed set of 40 hand-written probes includes the
cheese example above, which a user reported against an earlier model. Dev and test
remain pure PIZZA, so $A$ and $S$ keep their meaning.

\section{Six Encodings of One Tree}
\label{sec:formats}

\begin{table}[t]
\centering
\footnotesize
\setlength{\tabcolsep}{3pt}
\resizebox{\columnwidth}{!}{\begin{tabular}{@{}lrl@{}}
\toprule
Format & Tok. & Target \\
\midrule
\fmt{EXR} & 38 & \texttt{(ORDER (PIZZAORDER (NUMBER 1 ) (COM...} \\
\fmt{JSON} & 33 & \texttt{\{\textquotedbl{}orders\textquotedbl{}:[\{\textquotedbl{}type\textquotedbl{}:\textquotedbl{}p\textquotedbl{},\textquotedbl{}attrs\textquotedbl{}:[\{\textquotedbl{}n...} \\
\fmt{EXR-short} & 25 & \texttt{(O (P (N 1) (C (Q EXTRA) (T CHEESE)...} \\
\fmt{TAGGED} & 13 & \texttt{pizza n=1 extra=cheese topp=pepperoni} \\
\fmt{DSL} & 8 & \texttt{p 1 +cheese pepperoni} \\
\fmt{WORDS} & 7 & \texttt{pizza 1 extra cheese pepperoni} \\
\bottomrule
\end{tabular}
}
\caption{One test order, \emph{``\exUtterance''}, in the six encodings (long
targets truncated). Tokens are Qwen's.}
\label{tab:formats}
\end{table}

Comparing unrelated formats confounds length with form, so we built six encodings
of the \emph{identical} tree that decode back to it on every dev and test utterance
and on a 50{,}000-row training sample (Table~\ref{tab:formats}). \fmt{EXR} is the
dataset's s-expression. \fmt{JSON} is a conventional encoding of about the same
length: the matched-length control for \fmt{EXR}. \fmt{EXR-short} renames
\fmt{EXR}'s constructors to single letters. \fmt{TAGGED} is flat
\code{key=value} pairs. \fmt{DSL} is a terse grammar with sigils (\code{+} for an
extra quantity, \code{!} for negation), and \fmt{WORDS} has the same structure as
\fmt{DSL} with the markers written as ordinary words; it is \fmt{DSL}'s
matched-length control. \fmt{EXR}, \fmt{EXR-short}, \fmt{TAGGED} and \fmt{DSL}
form a length ladder.

\begin{figure}[t]
\centering
\includegraphics[width=\columnwidth]{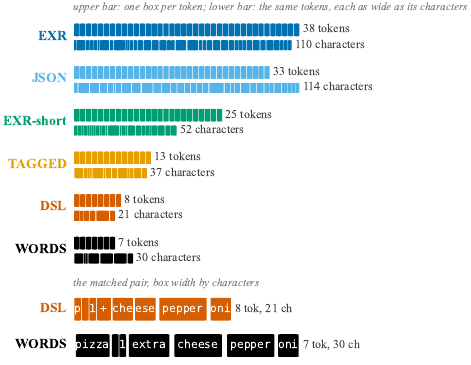}
\caption{The order of Table~\ref{tab:formats} under Qwen's tokenizer. For each
format the upper bar has one box per token and the lower bar draws the same tokens
as wide as their characters (both bars to a common scale across formats). Bottom:
the matched pair with its tokens; \fmt{WORDS} is longer as text and shorter in
tokens. Boundaries come from the tokenizer's own merge table.}
\label{fig:tok}
\end{figure}

\paragraph{Characters are not tokens.}
Over the \nTokTest{} gold test trees, \fmt{WORDS} is the longer of the matched
pair in characters (\chWords{} vs.\ \chDsl{}) but costs fewer Qwen tokens
(\tqWords{} vs.\ \tqDsl{}) and about the same number of LFM2 tokens (\tlWords{}
vs.\ \tlDsl{}; Appendix Table~\ref{tab:chars}). Figure~\ref{fig:tok} shows why:
\code{+cheese} splits into three subwords where \code{extra cheese} stays two.
Symbols save characters, not tokens. LFM2's vocabulary is smaller and inflates
the punctuation-heavy formats most (\fmt{JSON} \tqJson{}$\to$\tlJson{} tokens,
\fmt{WORDS} \tqWords{}$\to$\tlWords{}).

\section{Setup and Statistics}
\label{sec:setup}

\paragraph{Models and training.}
\Qlab{}~\citep{qwen} and \Llab{}~\citep{lfm}, a largely convolutional model with a
vocabulary about four times smaller. LoRA~\citep{lora} with rank 16 and
$\alpha{=}16$ on all linear projections, batch 16, cosine schedule, loss on the
completion only. Rank is held fixed across formats because the comparison is
between formats, not adapters; \citet{lora} found far lower ranks sufficient on
their tasks. Training runs to at most \maxSteps{} steps (\samplesSeen{} examples)
with early stopping every 100 steps (patience 10) on the mean of three dev
scores: exact match, refusal of dev negatives, and exact match on dev corrections.
The learning rate is chosen per model and format from $\{10^{-4}, 2{\cdot}10^{-4},
5{\cdot}10^{-4}\}$ by dev exact match at a pool of 100k over \nSweepSeeds{} seeds,
then frozen (Appendix Table~\ref{tab:lr}). The token cap per format is 1.25 times
its longest target, so no correct output can be truncated.

\paragraph{Grid.}
Six formats $\times$ pool sizes \{1k, 10k, 100k, 1M, \nHead{}\} $\times$
\nSeeds{} seeds $\times$ two models, \nGrid{} runs, every one with the refusal and
correction data. A seed fixes the training subset, the negatives and the
corrections, so formats are paired by seed. Since no run sees more than
\samplesSeen{} examples, the larger sizes differ in the diversity of the pool a
run samples from, not in examples seen; we call the axis \emph{pool size}.

\paragraph{Metrics.}
\label{sec:metrics}
Every test order ends in exactly one of four outcomes:
\[
\underbrace{A}_{\text{correct}} + \underbrace{S}_{\text{silent}}
+ \underbrace{D_\text{parse}}_{\text{unparseable}}
+ \underbrace{D_\text{refuse}}_{\text{refused}} = 1 .
\]
$A$ is exact match under the dataset's own matcher. $S$ is output that parses
into a tree that does not match: the caller cannot detect it. $D_\text{parse}$
does not parse, and $D_\text{refuse}$ is a refusal of a real order; both are
visible to the caller. We read them in this order: $S$ is the deployment
objective, because those errors reach the customer; $A$ is reported beside it for
comparison with prior work; and $D_\text{refuse}$ is the price of the guardrail.
Every table shows $A$ and $S$ together.

\paragraph{Statistics.}
Fixed before any result was read. A cell is the mean of \nSeeds{} seeds. Intervals
and $p$-values come from a hierarchical bootstrap (\nBoot{} resamples) over seeds
(paired by seed index across formats) and test utterances (paired across
formats). A comparison \emph{separates} when its Holm-adjusted $p<0.05$, the
family being the five pool sizes of one comparison on one model. Retraining one
cell twice moved test exact match by \Qrerun{} points on Qwen and \Lrerun{} on
LFM2, against a seed SD in that cell of \QrerunSeedSd{} and \LrerunSeedSd{}; we
therefore use the seed spread and the bootstrap, not a single rerun, as the
measure of noise.

\section{Results}
\label{sec:results}

\begin{table*}[t]
\centering
\small
\setlength{\tabcolsep}{5pt}
\begin{tabular}{lrrrrrrr}
\toprule
Format & Tok. & $A$ (EM) $\uparrow$ & $S$ (silent) $\downarrow$ & $D_\text{parse}$ & $D_\text{refuse}$ & Inv./utt. & ms \\
\midrule
\multicolumn{8}{l}{\textit{Qwen3.5-0.8B}} \\
\fmt{EXR} & 44.3 & \underline{85.8}\,{\scriptsize$\pm$1.6} & 10.3\,{\scriptsize$\pm$0.8} & 0.6\,{\scriptsize$\pm$0.8} & 3.3\,{\scriptsize$\pm$1.1} & 0.100 & 1171 \\
\fmt{JSON} & 44.1 & 81.6\,{\scriptsize$\pm$1.3} & 14.9\,{\scriptsize$\pm$1.3} & 0.3\,{\scriptsize$\pm$0.3} & 3.2\,{\scriptsize$\pm$1.0} & 0.126 & 1220 \\
\fmt{EXR-short} & 32.3 & 75.2\,{\scriptsize$\pm$3.6} & 17.4\,{\scriptsize$\pm$4.1} & 3.3\,{\scriptsize$\pm$2.0} & 4.2\,{\scriptsize$\pm$2.2} & 0.140 & 903 \\
\fmt{TAGGED} & 22.2 & \underline{85.9}\,{\scriptsize$\pm$2.9} & 11.6\,{\scriptsize$\pm$2.4} & 0.5\,{\scriptsize$\pm$0.3} & 2.1\,{\scriptsize$\pm$0.8} & 0.095 & 616 \\
\fmt{DSL} & 11.3 & 79.7\,{\scriptsize$\pm$1.9} & 8.3\,{\scriptsize$\pm$2.3} & 6.7\,{\scriptsize$\pm$3.2} & 5.4\,{\scriptsize$\pm$2.3} & 0.070 & 325 \\
\fmt{WORDS} & 10.9 & \textbf{87.5}\,{\scriptsize$\pm$0.8} & \textbf{4.5}\,{\scriptsize$\pm$1.5} & 4.6\,{\scriptsize$\pm$0.8} & 3.4\,{\scriptsize$\pm$1.8} & 0.018 & 317 \\
\midrule
\multicolumn{8}{l}{\textit{LFM2.5-350M}} \\
\fmt{EXR} & 52.4 & \underline{74.0}\,{\scriptsize$\pm$2.5} & 17.9\,{\scriptsize$\pm$2.8} & 0.2\,{\scriptsize$\pm$0.2} & 7.8\,{\scriptsize$\pm$4.2} & 0.123 & 821 \\
\fmt{JSON} & 57.8 & 70.7\,{\scriptsize$\pm$3.6} & 19.1\,{\scriptsize$\pm$1.2} & 0.1\,{\scriptsize$\pm$0.2} & 10.1\,{\scriptsize$\pm$3.3} & 0.135 & 926 \\
\fmt{EXR-short} & 35.6 & 70.8\,{\scriptsize$\pm$0.7} & 16.9\,{\scriptsize$\pm$2.6} & 2.1\,{\scriptsize$\pm$1.7} & 10.1\,{\scriptsize$\pm$2.3} & 0.141 & 571 \\
\fmt{TAGGED} & 24.4 & 71.6\,{\scriptsize$\pm$2.5} & 21.4\,{\scriptsize$\pm$3.8} & 0.7\,{\scriptsize$\pm$0.4} & 6.3\,{\scriptsize$\pm$3.9} & 0.150 & 395 \\
\fmt{DSL} & 11.4 & 72.0\,{\scriptsize$\pm$3.0} & \underline{13.1}\,{\scriptsize$\pm$2.6} & 5.7\,{\scriptsize$\pm$1.5} & 9.2\,{\scriptsize$\pm$3.5} & 0.086 & 178 \\
\fmt{WORDS} & 12.0 & \textbf{77.4}\,{\scriptsize$\pm$3.2} & \textbf{10.0}\,{\scriptsize$\pm$1.6} & 5.8\,{\scriptsize$\pm$0.8} & 6.7\,{\scriptsize$\pm$3.0} & 0.051 & 186 \\
\bottomrule
\end{tabular}

\caption{Test results at the full pool (\nHead{}), mean $\pm$ seed SD over
\nSeeds{} seeds, in percent of the \nTest{} human-written test orders. $A$ + $S$ +
$D_\text{parse}$ + $D_\text{refuse}$ = 100. Bold: best format in the column;
underlined: not separable from the best (Holm, $p\geq0.05$). Tok.: mean output
tokens. Inv./utt.: invented entities per order. ms: median latency at batch size
one on one GPU, mean over seeds.}
\label{tab:main}
\end{table*}

Table~\ref{tab:main} gives the full-pool results. On both models \fmt{WORDS} has
the highest exact match (\QwordsEm{}\% on Qwen; \QtiedEm{} not separable from it)
and the lowest silent rate (no other format separable on Qwen; \LtiedS{} on LFM2),
with the fewest tokens on Qwen and nearly the fewest on LFM2. For reference, the
published \publishedSystem{} system reaches \publishedEm{}\% on this test set with
the full pool \citep{pizza}; it is a fully fine-tuned encoder-decoder trained
without refusals or corrections, so we do not compare further.

\begin{figure}[t]
\centering
\includegraphics[width=\columnwidth]{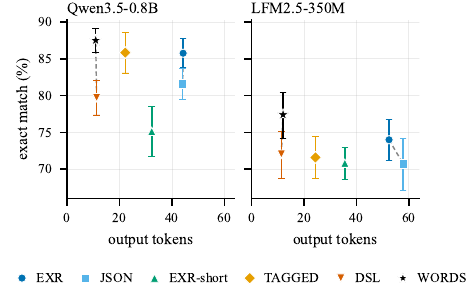}
\caption{Exact match against output tokens at the full pool, 95\% bootstrap
intervals. Dashed lines join the matched-length pairs (\fmt{EXR}/\fmt{JSON},
\fmt{DSL}/\fmt{WORDS}).}
\label{fig:acc}
\end{figure}

\subsection{Shorter is not better}

Regressing exact match on output tokens across the four ladder formats gives a
positive slope (removing tokens lowers accuracy) in \slopePos{} of \slopeCells{}
model$\times$size cells, never more than \slopeMax{} points per ten tokens, and it
separates in only \slopeSep{}. The ladder is not monotone (Figure~\ref{fig:acc}):
\fmt{EXR-short}, which only renames \fmt{EXR}'s constructors, loses
\QexrExrShortGapAbs{} points on Qwen, separating at every size, and \LexrExrShortGapAbs{}
on LFM2; \fmt{TAGGED}, half as long as \fmt{EXR}, matches it on Qwen. Length does
not predict accuracy in either direction.

\subsection{Surface form, not length}
\label{sec:form}

\begin{table}[t]
\centering
\footnotesize
\setlength{\tabcolsep}{3.2pt}
\resizebox{\columnwidth}{!}{\begin{tabular}{lrrrrr}
\toprule
Difference (points) & 1k & 10k & 100k & 1M & 2.5M \\
\midrule
\multicolumn{6}{l}{\textit{Qwen3.5-0.8B}} \\
\fmt{WORDS}$-$\fmt{DSL} & +6.7$^\dagger$ & +4.8$^\dagger$ & +6.2$^\dagger$ & +5.8$^\dagger$ & +7.8$^\dagger$ \\
\fmt{EXR}$-$\fmt{JSON} & +9.3$^\dagger$ & +3.8 & +3.6 & +5.5$^\dagger$ & +4.2$^\dagger$ \\
\fmt{EXR}$-$\fmt{EXR-short} & +6.2$^\dagger$ & +11.6$^\dagger$ & +10.3$^\dagger$ & +9.3$^\dagger$ & +10.6$^\dagger$ \\
\fmt{WORDS}$-$\fmt{EXR} & +4.0 & +1.8 & +1.6 & +0.0 & +1.8 \\
$S$: \fmt{EXR}$-$\fmt{WORDS} & +11.4$^\dagger$ & +5.1$^\dagger$ & +4.4$^\dagger$ & +5.0$^\dagger$ & +5.8$^\dagger$ \\
\midrule
\multicolumn{6}{l}{\textit{LFM2.5-350M}} \\
\fmt{WORDS}$-$\fmt{DSL} & +8.9$^\dagger$ & +6.7$^\dagger$ & +4.3$^\dagger$ & +4.3 & +5.4$^\dagger$ \\
\fmt{EXR}$-$\fmt{JSON} & +3.4 & +1.5 & +3.4 & +1.5 & +3.3 \\
\fmt{EXR}$-$\fmt{EXR-short} & -0.9 & +2.3 & +1.4 & +1.1 & +3.2$^\dagger$ \\
\fmt{WORDS}$-$\fmt{EXR} & +7.2$^\dagger$ & +4.9$^\dagger$ & +3.5 & +4.4$^\dagger$ & +3.4 \\
$S$: \fmt{EXR}$-$\fmt{WORDS} & +11.2$^\dagger$ & +8.6$^\dagger$ & +7.7$^\dagger$ & +8.7$^\dagger$ & +7.9$^\dagger$ \\
\bottomrule
\end{tabular}
}
\caption{Paired differences in exact match (points; last row: silent rate, positive
= \fmt{WORDS} lower) at every pool size, both models. $^\dagger$: separates (Holm
within the row). 95\% intervals in Appendix Table~\ref{tab:pairsci}.}
\label{tab:pairs}
\end{table}

The matched pairs separate the two explanations (Table~\ref{tab:pairs}).
\fmt{WORDS} and \fmt{DSL} share their structure and differ by less than a token;
they differ only in whether a marker is \code{+cheese} or \code{extra cheese}.
\fmt{WORDS} is ahead in every one of the \pairCells{} cells, by \QwordsDslGap{}
points on Qwen [\QwordsDslLo{}, \QwordsDslHi{}] and \LwordsDslGap{} on LFM2 at the
full pool, and separates in all but one. On LFM2 the pair's token order inverts:
\fmt{WORDS} emits \LwordsTok{} tokens against \fmt{DSL}'s \LdslTok{} and is still
the more accurate. A format that is longer and more accurate is not winning by
being short.

\fmt{EXR} and \fmt{JSON} are both bracket-heavy notations, so this pair varies
little along the natural-word axis. \fmt{EXR} is ahead in all \pairCells{} cells
but separates in only \QexrJsonSep{} (all on Qwen): a consistent but small effect,
as expected if what matters is words rather than brackets. Against the native
format, \fmt{WORDS} is ahead on both models (\QwordsExrGap{} and \LwordsExrGap{}
points at the full pool), separating in some sizes on LFM2 and none on Qwen; we
read this as matching \fmt{EXR}'s accuracy with \QtokRatio{}$\times$ fewer tokens,
not as beating it.

\subsection{Silent errors and where they come from}
\label{sec:silent}

\begin{figure}[t]
\centering
\includegraphics[width=\columnwidth]{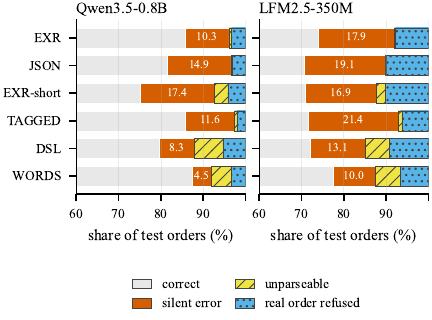}
\caption{Where the test orders go at the full pool (axis starts at 60\%). Labels:
silent error rate $S$.}
\label{fig:errors}
\end{figure}

Exact match and silent rate rank formats differently (Figure~\ref{fig:errors}).
\fmt{EXR} and \fmt{JSON} almost always parse, so nearly everything they get wrong
they get wrong silently. \fmt{WORDS} and \fmt{DSL} fail to parse more often, and
those failures are visible. \fmt{WORDS} has the lowest $S$: \QwordsS{}\% against
\QexrS{}\% for \fmt{EXR} on Qwen, a ratio of \QsilentRatio{} [\QsilentRatioLo{},
\QsilentRatioHi{}], and \LwordsS{}\% against \LexrS{}\% on LFM2, ratio
\LsilentRatio{} [\LsilentRatioLo{}, \LsilentRatioHi{}]. The difference separates
at every pool size on both models (Table~\ref{tab:pairs}, last rows). \fmt{WORDS}
also invents fewer entities that the customer never said: \QhallRatio{}$\times$
fewer than \fmt{EXR} on Qwen and \LhallRatio{}$\times$ on LFM2.

\begin{figure}[t]
\centering
\includegraphics[width=\columnwidth]{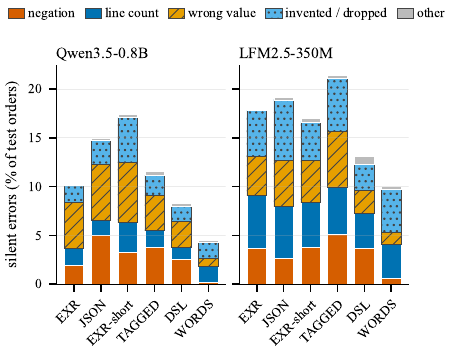}
\caption{Silent errors by kind at the full pool, as a share of all test orders
(bar height = $S$). Negation: a \code{NOT} attribute added, lost or misplaced;
line count: wrong number of pizzas or drinks.}
\label{fig:kinds}
\end{figure}

Classifying every silent error by what it changes (Figure~\ref{fig:kinds};
Appendix Table~\ref{tab:errors}) shows where the difference comes from. Negation is
\QwordsNegErr{}\% of \fmt{WORDS}'s silent errors on Qwen and \LwordsNegErr{}\% on
LFM2, against \QotherNegErrMin{}--\QotherNegErrMax{}\% and
\LotherNegErrMin{}--\LotherNegErrMax{}\% for the other formats; as a rate, Qwen
makes a silent negation error on \QexrNegRate{}\% of orders with \fmt{EXR} and
\QwordsNegRate{}\% with \fmt{WORDS}. \fmt{WORDS} writes \code{no onions} the way
customers say it, where the other formats translate it into a
\code{NOT}-construct or a sigil. What \fmt{WORDS} still gets wrong is mostly the
number of lines (\QwordsLineErr{}\% of its silent errors on Qwen).

\subsection{The guardrail: refusing and listening}
\label{sec:guard}

\begin{table*}[t]
\centering
\small
\setlength{\tabcolsep}{6pt}
\begin{tabular}{lrrrrr}
\toprule
Format & Neg.\ handled $\uparrow$ & Real refused $\downarrow$ & Corr.\ EM $\uparrow$ & Corr.\ applied & Probes /40 \\
\midrule
\multicolumn{6}{l}{\textit{Qwen3.5-0.8B}} \\
\fmt{EXR} & 99.0 & 3.3 & \underline{74.5}\,{\scriptsize$\pm$4.3} & 86.5 & 36.0 \\
\fmt{JSON} & 98.0 & 3.2 & 70.0\,{\scriptsize$\pm$2.8} & 85.7 & 37.8 \\
\fmt{EXR-short} & 96.1 & 4.2 & 62.9\,{\scriptsize$\pm$6.9} & 82.8 & 35.4 \\
\fmt{TAGGED} & 95.9 & 2.1 & \underline{73.0}\,{\scriptsize$\pm$4.5} & 87.7 & 37.2 \\
\fmt{DSL} & 98.9 & 5.4 & \underline{71.4}\,{\scriptsize$\pm$4.8} & 80.4 & 37.8 \\
\fmt{WORDS} & 98.6 & 3.4 & \textbf{75.3}\,{\scriptsize$\pm$4.7} & 85.0 & 38.4 \\
\midrule
\multicolumn{6}{l}{\textit{LFM2.5-350M}} \\
\fmt{EXR} & 98.2 & 7.8 & 55.5\,{\scriptsize$\pm$4.1} & 77.0 & 36.0 \\
\fmt{JSON} & 97.6 & 10.1 & 49.2\,{\scriptsize$\pm$10.7} & 72.1 & 34.2 \\
\fmt{EXR-short} & 98.1 & 10.1 & 53.4\,{\scriptsize$\pm$4.3} & 76.1 & 35.0 \\
\fmt{TAGGED} & 97.2 & 6.3 & 52.4\,{\scriptsize$\pm$8.6} & 77.8 & 33.6 \\
\fmt{DSL} & 98.3 & 9.2 & 54.6\,{\scriptsize$\pm$4.0} & 73.3 & 36.4 \\
\fmt{WORDS} & 98.2 & 6.7 & \textbf{62.6}\,{\scriptsize$\pm$3.0} & 77.7 & 36.2 \\
\bottomrule
\end{tabular}

\caption{Guardrail at the full pool, percent, mean of \nSeeds{} seeds. Neg.\
handled: held-out negatives refused (\nNeg{} per run). Real refused: test orders
refused ($D_\text{refuse}$). Corr.: \nCor{} test orders with held-out correction
cues; applied: the corrected attribute is right. Probes: hand-written corrections
correct, of 40. Bold/underline as in Table~\ref{tab:main}.}
\label{tab:guard}
\end{table*}

\paragraph{Refusal works for every format.}
Held-out negatives, built from vocabulary never seen in training, are refused
\QnegMin{}--\QnegMax{}\% of the time on Qwen and \LnegMin{}--\LnegMax{}\% on LFM2
(Table~\ref{tab:guard}). Chatter, gibberish and long text are refused essentially
always; the hardest kind is a real order with one unknown word, refused
\QunkMin{}--\QunkMax{}\% of the time on Qwen (Appendix Table~\ref{tab:negkinds}).
Format barely matters here.

\paragraph{The price is refusing real orders.}
\QrefuseMin{}--\QrefuseMax{}\% of real test orders are refused on Qwen and
\LrefuseMin{}--\LrefuseMax{}\% on LFM2. Reading them, they are genuine orders in
human phrasing the synthetic grammar never produces: greetings, thanks, and
stories (\emph{``i'm having a bunch of people over tonight so i need five
pizzas\ldots''}). Almost every refused order contains a word the training grammar never uses (\QoovRefMin{}--\QoovRefMax{}\% on Qwen, \LoovRefMin{}--\LoovRefMax{}\% on LFM2), against \QoovAns{}\% of the orders answered. An order with a greeting, \emph{please}
or \emph{thanks} (\nPolite{} of the test orders) is refused \QwordsPolite{}\% of
the time by Qwen \fmt{WORDS} against \QwordsNotPolite{}\% for other orders. The
strict rule meets politeness the training grammar does not contain.

\paragraph{Corrections are learned, and best by \fmt{WORDS}.}
On test orders with held-out cue phrases, exact match is
\QfixMin{}--\QfixMax{}\% on Qwen and \LfixMin{}--\LfixMax{}\% on LFM2, and the
correction itself is applied in \QappliedMin{}--\QappliedMax{}\% and
\LappliedMin{}--\LappliedMax{}\% of cases. \fmt{WORDS} is the best format on both
models, separating from every other format on LFM2. The reported cheese failure is
fixed in \probeFixed{} of \probeRuns{} full-pool runs. Training cues score
\QcueGapMin{}--\QcueGapMax{} points above held-out cues on Qwen, so part of what is
learned is the phrasing of the cue.

\subsection{Pool size}

\begin{figure}[t]
\centering
\includegraphics[width=\columnwidth]{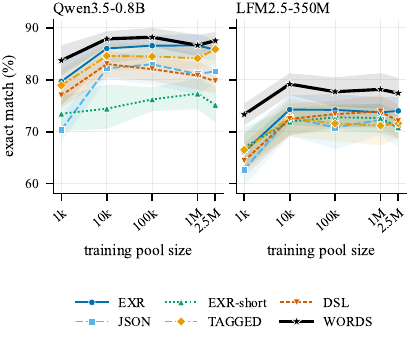}
\caption{Exact match by training pool size, 95\% bootstrap bands.}
\label{fig:pool}
\end{figure}

\fmt{WORDS} is the most accurate format at every pool size on both models
(Figure~\ref{fig:pool}). Its lead over \fmt{EXR} is largest at 1k
(\QwordsExrGapSmall{} points on Qwen, \LwordsExrGapSmall{} on LFM2) and narrows
with pool size; its lead over \fmt{DSL} does not narrow on Qwen
(\QwordsDslGapSmall{} at 1k, \QwordsDslGap{} at the full pool). Natural words help
most when data is scarce, but the advantage over a symbolic form of the same
length does not go away.

\subsection{Latency}

Median latency at batch size one is linear in output tokens (Qwen: \QmsFixed{}\,ms
+ \QmsPerTok{}\,ms per token, $R^2{=}\QlatRsq{}$; Appendix
Figure~\ref{fig:latency}) with throughput flat across formats, so \fmt{WORDS}
answers \Qspeedup{}$\times$ faster than \fmt{EXR} on Qwen and \Lspeedup{}$\times$
on LFM2. The process's GPU memory moves only between \QmemMin{} and
\QmemMax{}\,MiB on Qwen: format buys latency and accuracy, not memory.

\section{Discussion}

\paragraph{Which metric, when.}
If a parser's output is acted on without review, $S$ is the number that matters:
a format with higher $A$ but higher $S$ ships more wrong orders. $A$ is the right
number to compare with prior work, and $D_\text{parse}+D_\text{refuse}$ is what a
retry or a human fallback has to absorb. On PIZZA the three point the same way for
\fmt{WORDS}, which need not hold elsewhere; reporting all four outcomes makes the
trade visible.

\paragraph{What transfers.}
The recipe is to compress the structure of the target and keep its vocabulary
natural. It costs nothing at inference, needs no new labels, and is measurable on
any parser with the four-way split. Whether the gap persists for larger models and
other tasks is open (Limitations).

\paragraph{Strict refusal.}
Refusing anything that is not purely an order keeps the consumer from trusting
half a ticket, at the cost of \QrefuseMin{}--\LrefuseMax{}\% of real orders. A
softer rule, or negatives written in human rather than grammar style, could lower
that cost; we leave it to future work and report the strict rule's price as is.

\section{Conclusion}

Output format is a free design variable for a fine-tuned parser. On PIZZA, with two
model families, five pool sizes and five seeds, writing the target in plain words
matches or beats the native format's accuracy with a quarter of the tokens and
roughly halves the errors nobody can detect, mostly by getting negation right. The
gain survives a refusal and self-correction guardrail trained into every run, and a
tokenizer under which the word form is the longer one.

\section*{Limitations}

\emph{One dataset.} All results are on PIZZA, whose training pool comes from a
grammar with a small vocabulary; the human-written test set is the main guard
against learning the grammar alone. \emph{Small models.} Both models are under a
billion parameters; the effect may shrink with scale. \emph{Author-generated
guardrail data.} Negatives, corrections and probes are generated or written by us,
and the human test set contains neither; \cueRows{} test orders (2.4\%) use
\emph{make it}, \emph{make that} or \emph{hold on} in PIZZA's own sense, which our
generator teaches as correction cues, bounding any effect on test exact match at
\cueBound{} points. Every run includes the guardrail, so this campaign does not
measure what the guardrail costs in accuracy. \emph{Tuning.} Three
learning rates were at the edge of the grid (Appendix Table~\ref{tab:lr}); rank was
not tuned; \lateRuns{} of \nGrid{} runs (all LFM2) reached their best checkpoint
within 100 steps of the step limit. \emph{Generation.} \capHits{} of
\nGenerations{} generations hit the token cap, all runaway outputs longer than any
correct ticket. \emph{Deployment.} The exported LFM2 GGUF agrees with its trained
run on \lfmGgufAgree{}\% of orders; no number in this paper comes from a GGUF.

\section*{Ethics Statement}

PIZZA is released under CC-BY-NC-4.0 and contains no personal data; we
redistribute no dataset text. All runs used two NVIDIA RTX 6000 Ada GPUs.
A parser that refuses real orders shifts work to staff or customers; we report
that rate rather than hide it.

\bibliography{refs}

\appendix

\section{Additional Results}

\begin{figure}[t]
\centering
\includegraphics[width=\columnwidth]{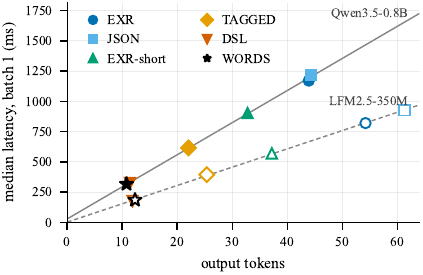}
\caption{Median latency at batch size one against output tokens, mean over five
seeds; filled: Qwen, hollow: LFM2. Lines: least-squares fits.}
\label{fig:latency}
\end{figure}

Tables~\ref{tab:sizes} to \ref{tab:errors} and Figure~\ref{fig:latency} give the
numbers behind Section~\ref{sec:results} at every pool size and by kind. All are
generated from the run outputs by the scripts in \artifacts.

\begin{table*}[t]
\centering
\scriptsize
\setlength{\tabcolsep}{3pt}
\begin{tabular}{lrrrrrrrrrr}
\toprule
 & \multicolumn{2}{c}{1k} & \multicolumn{2}{c}{10k} & \multicolumn{2}{c}{100k} & \multicolumn{2}{c}{1M} & \multicolumn{2}{c}{2.5M} \\
 & $A$ & $S$ & $A$ & $S$ & $A$ & $S$ & $A$ & $S$ & $A$ & $S$ \\
\midrule
\multicolumn{11}{l}{\textit{Qwen3.5-0.8B}} \\
\fmt{EXR} & 79.7{\tiny$\pm$4.9} & 17.0 & 86.0{\tiny$\pm$2.7} & 9.6 & 86.6{\tiny$\pm$2.0} & 9.2 & 86.6{\tiny$\pm$2.1} & 10.1 & 85.8{\tiny$\pm$1.6} & 10.3 \\
\fmt{JSON} & 70.4{\tiny$\pm$1.4} & 25.1 & 82.2{\tiny$\pm$1.3} & 14.6 & 83.0{\tiny$\pm$4.4} & 12.9 & 81.1{\tiny$\pm$4.6} & 11.5 & 81.6{\tiny$\pm$1.3} & 14.9 \\
\fmt{EXR-short} & 73.5{\tiny$\pm$3.5} & 18.8 & 74.4{\tiny$\pm$4.9} & 15.0 & 76.2{\tiny$\pm$1.4} & 15.1 & 77.3{\tiny$\pm$2.9} & 13.4 & 75.2{\tiny$\pm$3.6} & 17.4 \\
\fmt{TAGGED} & 78.9{\tiny$\pm$3.7} & 16.8 & 84.6{\tiny$\pm$2.9} & 12.1 & 84.5{\tiny$\pm$2.5} & 10.8 & 84.1{\tiny$\pm$1.1} & 12.4 & 85.9{\tiny$\pm$2.9} & 11.6 \\
\fmt{DSL} & 77.0{\tiny$\pm$1.9} & 10.5 & 83.0{\tiny$\pm$2.4} & 8.6 & 82.0{\tiny$\pm$1.9} & 8.9 & 80.8{\tiny$\pm$2.6} & 8.1 & 79.7{\tiny$\pm$1.9} & 8.3 \\
\fmt{WORDS} & 83.7{\tiny$\pm$3.1} & 5.6 & 87.9{\tiny$\pm$0.4} & 4.6 & 88.2{\tiny$\pm$1.3} & 4.8 & 86.6{\tiny$\pm$1.8} & 5.1 & 87.5{\tiny$\pm$0.8} & 4.5 \\
\midrule
\multicolumn{11}{l}{\textit{LFM2.5-350M}} \\
\fmt{EXR} & 66.1{\tiny$\pm$3.6} & 25.9 & 74.3{\tiny$\pm$2.5} & 18.6 & 74.2{\tiny$\pm$3.3} & 17.4 & 73.7{\tiny$\pm$2.4} & 17.4 & 74.0{\tiny$\pm$2.5} & 17.9 \\
\fmt{JSON} & 62.7{\tiny$\pm$3.0} & 29.4 & 72.7{\tiny$\pm$4.1} & 16.4 & 70.8{\tiny$\pm$4.3} & 18.3 & 72.3{\tiny$\pm$2.9} & 18.1 & 70.7{\tiny$\pm$3.6} & 19.1 \\
\fmt{EXR-short} & 67.0{\tiny$\pm$4.6} & 22.8 & 71.9{\tiny$\pm$1.9} & 18.6 & 72.8{\tiny$\pm$1.6} & 19.1 & 72.6{\tiny$\pm$2.4} & 17.6 & 70.8{\tiny$\pm$0.7} & 16.9 \\
\fmt{TAGGED} & 66.5{\tiny$\pm$2.7} & 24.0 & 72.7{\tiny$\pm$3.6} & 20.5 & 71.6{\tiny$\pm$3.5} & 18.8 & 71.2{\tiny$\pm$3.8} & 18.9 & 71.6{\tiny$\pm$2.5} & 21.4 \\
\fmt{DSL} & 64.4{\tiny$\pm$2.9} & 19.2 & 72.5{\tiny$\pm$2.8} & 15.2 & 73.4{\tiny$\pm$1.8} & 14.2 & 73.9{\tiny$\pm$2.9} & 13.6 & 72.0{\tiny$\pm$3.0} & 13.1 \\
\fmt{WORDS} & 73.3{\tiny$\pm$0.9} & 14.7 & 79.2{\tiny$\pm$1.3} & 10.1 & 77.7{\tiny$\pm$2.7} & 9.7 & 78.1{\tiny$\pm$3.5} & 8.7 & 77.4{\tiny$\pm$3.2} & 10.0 \\
\bottomrule
\end{tabular}

\caption{Exact match $A$ ($\pm$ seed SD) and silent rate $S$ at every pool size, percent.}
\label{tab:sizes}
\end{table*}

\begin{table*}[t]
\centering
\scriptsize
\setlength{\tabcolsep}{3pt}
\begin{tabular}{lrrrrr}
\toprule
Difference (points) & 1k & 10k & 100k & 1M & 2.5M \\
\midrule
\multicolumn{6}{l}{\textit{Qwen3.5-0.8B}} \\
\fmt{WORDS}$-$\fmt{DSL} & +6.7 [+3.4, +10.2]$^\dagger$ & +4.8 [+2.3, +7.3]$^\dagger$ & +6.2 [+3.9, +8.4]$^\dagger$ & +5.8 [+3.4, +8.6]$^\dagger$ & +7.8 [+6.1, +9.5]$^\dagger$ \\
\fmt{EXR}$-$\fmt{JSON} & +9.3 [+5.4, +12.8]$^\dagger$ & +3.8 [+0.2, +6.7] & +3.6 [-1.1, +7.8] & +5.5 [+2.2, +9.2]$^\dagger$ & +4.2 [+2.4, +5.8]$^\dagger$ \\
\fmt{EXR}$-$\fmt{EXR-short} & +6.2 [+3.3, +9.2]$^\dagger$ & +11.6 [+4.7, +16.6]$^\dagger$ & +10.3 [+7.9, +12.7]$^\dagger$ & +9.3 [+6.2, +12.5]$^\dagger$ & +10.6 [+7.9, +13.4]$^\dagger$ \\
\fmt{WORDS}$-$\fmt{EXR} & +4.0 [+0.5, +7.7] & +1.8 [-0.4, +4.4] & +1.6 [-0.8, +4.2] & +0.0 [-1.3, +1.3] & +1.8 [+0.1, +3.4] \\
$S$: \fmt{EXR}$-$\fmt{WORDS} & +11.4 [+8.9, +13.9]$^\dagger$ & +5.1 [+3.3, +7.1]$^\dagger$ & +4.4 [+2.8, +6.2]$^\dagger$ & +5.0 [+3.1, +7.1]$^\dagger$ & +5.8 [+4.3, +7.4]$^\dagger$ \\
\midrule
\multicolumn{6}{l}{\textit{LFM2.5-350M}} \\
\fmt{WORDS}$-$\fmt{DSL} & +8.9 [+5.5, +11.9]$^\dagger$ & +6.7 [+3.5, +9.9]$^\dagger$ & +4.3 [+2.2, +6.3]$^\dagger$ & +4.3 [-1.2, +8.9] & +5.4 [+3.3, +7.4]$^\dagger$ \\
\fmt{EXR}$-$\fmt{JSON} & +3.4 [-1.7, +8.5] & +1.5 [-3.1, +5.8] & +3.4 [-0.5, +9.0] & +1.5 [-1.4, +4.1] & +3.3 [-0.4, +7.3] \\
\fmt{EXR}$-$\fmt{EXR-short} & -0.9 [-4.8, +2.2] & +2.3 [-0.8, +5.5] & +1.4 [-2.2, +4.3] & +1.1 [-2.0, +5.1] & +3.2 [+0.9, +5.4]$^\dagger$ \\
\fmt{WORDS}$-$\fmt{EXR} & +7.2 [+3.7, +10.7]$^\dagger$ & +4.9 [+2.5, +7.5]$^\dagger$ & +3.5 [-0.0, +7.5] & +4.4 [+1.0, +7.8]$^\dagger$ & +3.4 [-0.9, +7.6] \\
$S$: \fmt{EXR}$-$\fmt{WORDS} & +11.2 [+8.6, +13.7]$^\dagger$ & +8.6 [+5.9, +11.4]$^\dagger$ & +7.7 [+4.3, +10.6]$^\dagger$ & +8.7 [+6.0, +11.5]$^\dagger$ & +7.9 [+4.2, +11.6]$^\dagger$ \\
\bottomrule
\end{tabular}

\caption{Table~\ref{tab:pairs} with 95\% hierarchical bootstrap intervals.}
\label{tab:pairsci}
\end{table*}

\begin{table}[t]
\centering
\footnotesize
\setlength{\tabcolsep}{3pt}
\begin{tabular}{lrrrrr}
\toprule
Format & Chars & \multicolumn{2}{c}{Qwen3.5} & \multicolumn{2}{c}{LFM2.5} \\
 & & tok. & ch./tok. & tok. & ch./tok. \\
\midrule
\fmt{EXR} & 123.0 & 45.0 & 2.7 & 58.0 & 2.1 \\
\fmt{JSON} & 150.7 & 45.6 & 3.3 & 64.7 & 2.3 \\
\fmt{EXR-short} & 66.0 & 33.4 & 2.0 & 39.0 & 1.7 \\
\fmt{TAGGED} & 63.4 & 21.7 & 2.9 & 25.7 & 2.5 \\
\fmt{DSL} & 34.9 & 10.8 & 3.2 & 11.7 & 3.0 \\
\fmt{WORDS} & 43.1 & 10.3 & 4.2 & 11.9 & 3.6 \\
\bottomrule
\end{tabular}

\caption{Mean characters and tokens of the gold targets over all \nTokTest{} test
orders, for both tokenizers.}
\label{tab:chars}
\end{table}

\begin{table}[t]
\centering
\footnotesize
\setlength{\tabcolsep}{3pt}
\begin{tabular}{lrrrl}
\toprule
Format & 1e-4 & 2e-4 & 5e-4 & \\
\midrule
\multicolumn{5}{l}{\textit{Qwen3.5-0.8B}} \\
\fmt{EXR} & 84.77 & \textbf{85.82} & 80.36 &  \\
\fmt{JSON} & \textbf{83.62} & 83.62 & 80.27 & edge \\
\fmt{EXR-short} & 76.82 & 79.89 & \textbf{79.98} & edge \\
\fmt{TAGGED} & 86.11 & \textbf{87.07} & 81.13 &  \\
\fmt{DSL} & 81.13 & \textbf{82.95} & 79.02 &  \\
\fmt{WORDS} & 86.11 & \textbf{86.78} & 79.50 &  \\
\midrule
\multicolumn{5}{l}{\textit{LFM2.5-350M}} \\
\fmt{EXR} & 69.83 & \textbf{75.57} & 70.50 &  \\
\fmt{JSON} & 69.54 & \textbf{70.02} & 69.54 &  \\
\fmt{EXR-short} & 71.46 & \textbf{75.77} & 66.86 &  \\
\fmt{TAGGED} & 74.62 & \textbf{75.00} & 70.21 &  \\
\fmt{DSL} & \textbf{75.57} & 72.70 & 67.05 & edge \\
\fmt{WORDS} & 78.93 & \textbf{79.02} & 69.83 &  \\
\bottomrule
\end{tabular}

\caption{Learning-rate selection: dev exact match at a 100k pool, mean of
\nSweepSeeds{} seeds; bold is the rate used. ``edge'': chosen at the end of the grid.}
\label{tab:lr}
\end{table}

\begin{table}[t]
\centering
\scriptsize
\setlength{\tabcolsep}{2.2pt}
\resizebox{\columnwidth}{!}{\begin{tabular}{lrrrrrrr}
\toprule
Format & junk & long & non-order & off-menu & mixed & unknown word & probe \\
\midrule
\multicolumn{8}{l}{\textit{Qwen3.5-0.8B}} \\
\fmt{EXR} & 100.0 & 100.0 & 99.4 & 99.8 & 99.3 & 96.1 & 100.0 \\
\fmt{JSON} & 100.0 & 100.0 & 100.0 & 98.8 & 98.3 & 92.9 & 100.0 \\
\fmt{EXR-short} & 99.7 & 100.0 & 99.4 & 97.5 & 96.3 & 87.0 & 100.0 \\
\fmt{TAGGED} & 100.0 & 100.0 & 99.1 & 95.9 & 96.9 & 86.3 & 100.0 \\
\fmt{DSL} & 100.0 & 100.0 & 99.4 & 99.8 & 99.0 & 96.0 & 100.0 \\
\fmt{WORDS} & 100.0 & 100.0 & 99.7 & 99.3 & 97.6 & 95.9 & 100.0 \\
\midrule
\multicolumn{8}{l}{\textit{LFM2.5-350M}} \\
\fmt{EXR} & 100.0 & 100.0 & 100.0 & 98.9 & 99.4 & 92.2 & 100.0 \\
\fmt{JSON} & 100.0 & 100.0 & 100.0 & 98.3 & 99.7 & 89.6 & 100.0 \\
\fmt{EXR-short} & 100.0 & 100.0 & 99.7 & 98.7 & 99.8 & 91.6 & 100.0 \\
\fmt{TAGGED} & 100.0 & 100.0 & 100.0 & 96.5 & 98.3 & 90.5 & 100.0 \\
\fmt{DSL} & 100.0 & 100.0 & 100.0 & 98.3 & 99.3 & 93.7 & 100.0 \\
\fmt{WORDS} & 100.0 & 100.0 & 100.0 & 98.4 & 99.2 & 93.0 & 100.0 \\
\bottomrule
\end{tabular}
}
\caption{Held-out negatives refused, by kind, full pool, percent.}
\label{tab:negkinds}
\end{table}

\begin{table*}[t]
\centering
\scriptsize
\begin{tabular}{lrrrrrrrrr}
\toprule
Format & swap & retract & only & size & number & drink & cancel & restart & decoy \\
\midrule
\multicolumn{10}{l}{\textit{Qwen3.5-0.8B}} \\
\fmt{EXR} & 73.1 & 73.2 & 82.3 & 81.7 & 86.7 & 59.6 & 61.8 & 78.6 & 74.4 \\
\fmt{JSON} & 66.8 & 63.5 & 85.6 & 75.7 & 80.4 & 56.9 & 66.0 & 77.8 & 64.9 \\
\fmt{EXR-short} & 60.3 & 58.0 & 79.7 & 72.6 & 74.9 & 46.3 & 57.6 & 61.0 & 59.6 \\
\fmt{TAGGED} & 72.3 & 67.2 & 84.7 & 75.2 & 82.7 & 63.8 & 67.6 & 79.2 & 70.1 \\
\fmt{DSL} & 72.9 & 72.1 & 80.3 & 72.1 & 78.9 & 62.8 & 67.8 & 75.6 & 63.3 \\
\fmt{WORDS} & 72.3 & 74.9 & 89.1 & 81.1 & 79.7 & 60.1 & 68.2 & 77.9 & 73.7 \\
\midrule
\multicolumn{10}{l}{\textit{LFM2.5-350M}} \\
\fmt{EXR} & 53.2 & 53.9 & 71.3 & 67.3 & 73.3 & 32.5 & 44.4 & 52.6 & 51.4 \\
\fmt{JSON} & 49.8 & 50.9 & 65.6 & 60.2 & 63.5 & 24.3 & 39.4 & 37.1 & 45.8 \\
\fmt{EXR-short} & 54.7 & 56.2 & 69.4 & 64.8 & 69.6 & 28.3 & 37.9 & 44.5 & 49.9 \\
\fmt{TAGGED} & 45.6 & 54.7 & 72.6 & 56.9 & 62.9 & 34.9 & 40.3 & 50.9 & 51.2 \\
\fmt{DSL} & 54.0 & 59.8 & 66.8 & 58.3 & 66.7 & 31.8 & 42.6 & 54.4 & 51.2 \\
\fmt{WORDS} & 60.0 & 65.7 & 70.9 & 76.2 & 70.9 & 43.1 & 51.6 & 53.8 & 63.4 \\
\bottomrule
\end{tabular}

\caption{Held-out corrections, exact match by kind, full pool, percent.}
\label{tab:corkinds}
\end{table*}

\begin{table*}[t]
\centering
\scriptsize
\begin{tabular}{lrrrrrrrr}
\toprule
Format & $n$ & line\ count & negation & topping\ swapped & value\ swapped & invented & dropped & other\ structure \\
\midrule
\multicolumn{9}{l}{\textit{Qwen3.5-0.8B}} \\
\fmt{EXR} & 698 & 17 & 19 & 15 & 31 & 9 & 9 & 1 \\
\fmt{JSON} & 1011 & 10 & 34 & 11 & 28 & 3 & 13 & 1 \\
\fmt{EXR-short} & 1180 & 18 & 19 & 18 & 18 & 8 & 18 & 2 \\
\fmt{TAGGED} & 786 & 14 & 33 & 11 & 21 & 4 & 14 & 4 \\
\fmt{DSL} & 561 & 15 & 30 & 12 & 20 & 7 & 13 & 3 \\
\fmt{WORDS} & 303 & 36 & 5 & 3 & 15 & 11 & 25 & 5 \\
\midrule
\multicolumn{9}{l}{\textit{LFM2.5-350M}} \\
\fmt{EXR} & 1216 & 30 & 20 & 7 & 15 & 4 & 22 & 1 \\
\fmt{JSON} & 1298 & 28 & 14 & 7 & 18 & 7 & 25 & 1 \\
\fmt{EXR-short} & 1150 & 27 & 22 & 9 & 17 & 7 & 16 & 2 \\
\fmt{TAGGED} & 1450 & 23 & 24 & 6 & 20 & 2 & 23 & 1 \\
\fmt{DSL} & 887 & 27 & 28 & 7 & 12 & 6 & 14 & 6 \\
\fmt{WORDS} & 681 & 35 & 6 & 4 & 9 & 19 & 24 & 4 \\
\bottomrule
\end{tabular}

\caption{Silent errors by kind, full pool, all seeds pooled: $n$ errors, and the
percent of them in each kind.}
\label{tab:errors}
\end{table*}

\section{Integrity Checks}

Before reading any result we checked: all \nGrid{} runs complete; exact match and
parse validity recomputed from the raw outputs with the dataset's matcher equal
the stored values in every run; negatives and corrections identical across
models, formats and sizes for a seed; no evaluation utterance in any training
subset; batched and batch-one generation agree on all but at most 2 of 200
utterances per run; left padding in every run. The analysis code and the checks
are in \artifacts.

\section{Deployment Demonstration}

The café parser that motivated this work, a separate model with its own compact
format, runs entirely on an iPad (\padDevice{}, A16) through llama.cpp with Metal
offload: Q8\_0 weights (795\,MiB file, \padMem{} resident), \padLoad{} load time,
median \padLatency{} and p90 \padPninety{} per order, \padTps{}. Latency on the
device is close to linear in output tokens, so the token savings of a compact
format translate directly into response time. This is a demonstration on ten
orders, not an evaluation.

\end{document}